\documentclass[journal]{IEEEtran}
\usepackage{hyperref}
\usepackage{color,soul}
\usepackage{graphicx} 
\usepackage{multirow}
\usepackage{tabularx}
\usepackage{booktabs}
\usepackage{float}
\usepackage{placeins}

\ifCLASSINFOpdf
\else
\fi
\begin{document}

\title{A High-Resolution UAV Benchmark Dataset and Baseline Evaluation for Teak Crown Segmentation in Western Ghats Tropical Forests}

\author{
  \textit{Anand~S~Sahadevan$^{*}$, Sunita~Arya$^{*}$, Anup~Kumar~Das, Bimal~Kumar~Bhattacharya and S~Manthira~Moorthi \\}
  \vspace{0.2cm}
  
  \large Space Applications Centre \\
  Indian Space Research Organisation (ISRO), India
  \thanks{*Equal Contribution}

}

%



\maketitle

\begin{abstract}
Accurate teak-tree segmentation from high-resolution drone imagery is critical for sustainable plantation management, yet lacks dedicated benchmarks. We introduce TeakTreeNet, a comprehensive dataset and evaluation protocol for binary teak-tree segmentation at native 2cm resolution. Four state-of-the-art architectures: U-Net, PSPNet, DeepLabV3, and SegFormer were rigorously benchmarked across scratch and ImageNet-pretrained settings with three-seed reproducibility. SegFormer establishes new SOTA performance with F1 0.938±0.002 (Teak), IoU 0.968±0.001 (Teak), and Accuracy 0.986±0.001  outperforming all pretrained CNNs. Domain-adaptive training suffices, as the scratch SegFormer outperforms ImageNet pretrained baselines, avoiding large-scale pretraining overhead. Comprehensive analysis reveals Transformer's superior boundary delineation for dense plantation scenarios. TeakTreeNet provides researchers with 6391 expertly annotated drone patches (224×224) and establishes reproducible evaluation protocols, enabling direct comparison of future methods. Our benchmark demonstrates SegFormer's optimal balance of accuracy, efficiency, and deployability for operational forestry monitoring of teak plantations for India's western tropical forests.
\end{abstract}


%
\IEEEpeerreviewmaketitle


\section{Introduction}
Teak (Tectona grandis) is one of the most valuable tropical hardwood species, renowned for its strength, durability, and natural resistance to pests. Accurate mapping and continuous monitoring of forest tree species is recognized as critical for sustainable management
\cite{fao2018state}. Traditional ground surveys, while reliable, are time consuming, costly, and often cover only limited areas. In recent years, high-resolution drone imagery has transformed the way teak resources are assessed. Drones offer detailed spatial information, flexible and frequent data collection, and a cost-effective alternative to field surveys. 
Despite rapid progress in remote sensing and computer vision, publicly available benchmark datasets for tree crown delineation remain geographically limited, with less representation of Indian tropical forests and species-specific crown segmentation tasks. In particular, manually annotated Unmanned Aerial Vehicle (UAV) datasets covering spectral red, blue, green (RGB) channels that support reproducible training, validation, and model comparison for teak crown delineation are limited in open data repositories. However, widely used resources such as the National Ecological Observation Network (NEON) Tree Evaluation benchmarks \cite{weinstein2021benchmark} provide standardized evaluation datasets for tree crown detection/delineation in multiple types of forests in the United States. Although there are UAV benchmark datasets, they are typically collected in non-Indian settings and forest compositions \cite{troles2024bamforests}, limiting direct transferability to structurally complex and species-rich forests of the Western Ghats. 
\begin{figure}[!t]
\centering
\includegraphics[width=0.48\textwidth]{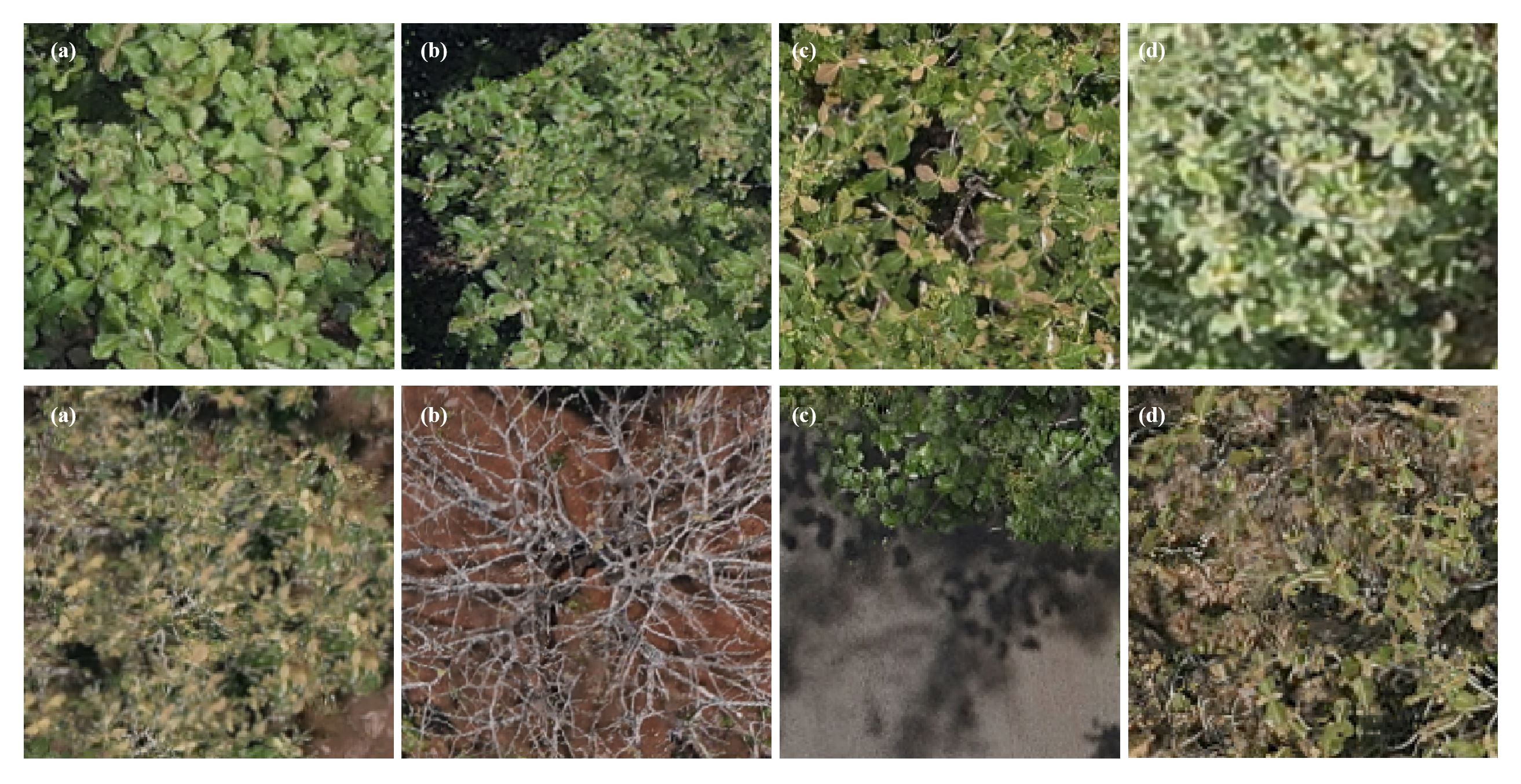}
\caption{Diverse sample image patches of teak and non-teak classes extracted from UAV-derived RGB orthomosaics. The top row shows patches of teak canopy, while the bottom row shows patches that are not classified as teak canopy}
\label{fig1}
\end{figure} 
Several open and widely adopted frameworks have accelerated crown delineation research by providing pretrained models and reproducible baselines, notably DeepForest \cite{weinstein2020deepforest}, which was designed to generalize crown delineation in RGB data via large-scale pretraining and fine-tuning. Recent tropical studies have demonstrated that deep learning can delineate tree crowns from aerial RGB imagery even in complex canopies, but performance remains sensitive to crown overlap, spatial resolution, and domain shift between regions \cite{duguay2026selvamask}. 
\begin{figure*}[!tbh]
\centering
\includegraphics[width=0.98\textwidth]{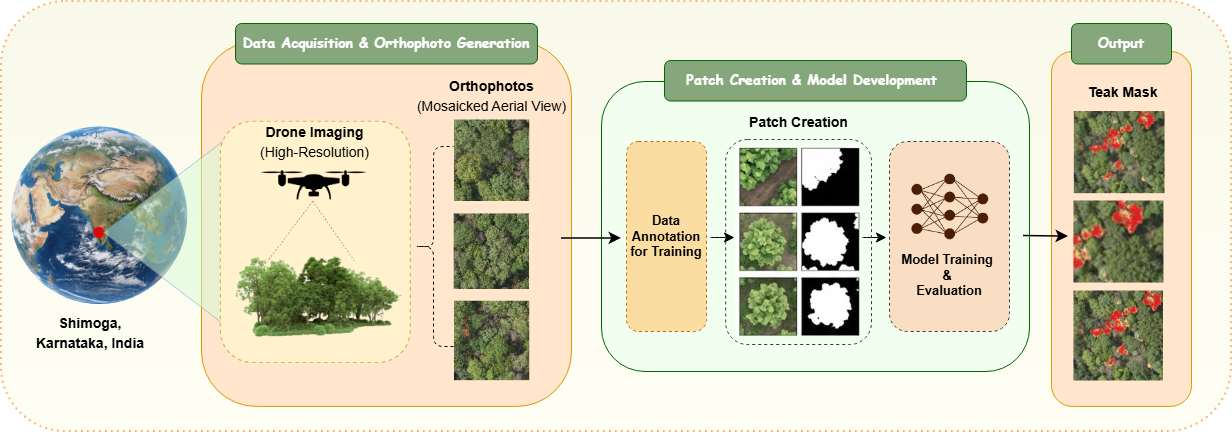}
\caption{An end-to-end workflow for teak-tree crown delineation using UAV-derived RGB orthomosaics.}
\label{fig2}
\end{figure*}

In this context, this work introduces a benchmark UAV RGB orthomosaic data set from the tropical forests of the western Ghats, India, with curated teak crown annotations designed specifically for reproducible evaluation of teak crown segmentation in natural forest conditions. The objectives are: (1) to develop a high-resolution drone RGB benchmark that captures the radiometric and structural complexity of teak crowns in mixed tropical canopies; (2) investigates whether transfer learning through large-scale pre-training can enhance teak crown delineation performance, particularly under limited annotated data and complex canopy conditions; and (3) to provide a rigorous comparison of representative state-of-the-art crown segmentation models under a consistent training/validation protocol and standardized metrics. Accordingly, this study aims to establish robust baseline performance metrics, identify model-specific failure modes in teak crown delineation, and provide a standardized and regionally representative benchmark dataset for teak crown delineation.

\section{Study Area and  Data}
\subsection{Study Area}
The Shimoga forests, Karnataka, India, and were chosen for developing benchmark data for this study. These study sites were established as part of the NASA-ISRO-synthetic-aperture-radar (NISAR) mission to support the calibration and validation of the above-ground-biomass (AGB) product \cite{das2024towards}. Tropical semi-evergreen and mixed deciduous forests vegetation dominated by teak plantations make it a significant site for this study. Larger plots of 1ha sizes were established based on Committee on Earth Observation Satellites (CEOS) protocol \cite{duncanson2021aboveground} on forest AGB mapping. In Shimoga, ten randomly located plots (and $3 \times 3$ sub-plots with in each plot) were established. Each square sized 1 hectare plot was laid in N/S and E/W directions for the principal axes of the plot. The tree species names, tree height, tree girth, stem diameter at breast height, and geo-locations were measured at each plot.

\subsection{RGB imaging using Drone}
Aerial imagery of the forest canopy over Shimoga was acquired (March, 2022) using a DJI Zenmuse P1 photogrammetric sensor integrated with a DJI Matrice 300 RTK UAV. The Zenmuse P1 is a high-performance RGB camera featuring a 45-megapixel full-frame sensor ($35.9 \times 24$mm). The system utilized the TimeSync 2.0 synchronization module to align the camera, flight controller, and RTK positioning at the microsecond level, ensuring centimeter-level absolute accuracy without the need for extensive ground control points (GCPs). Missions were conducted at a flight speed of 15 m/s with a 75\% front and 55\% side overlap, resulting in a measured Ground Sample Distance (GSD) of 2cm. These raw images are then processed using pix4D photogrammetric software to create a georeferenced orthomosaic map of each plot. 

Teak is characterized by a distinct set of biophysical and phenological features that facilitate its identification in high-resolution RGB images captured by drones, in contrast with many other tropical tree species. Its broad leaves with a rough texture, open crown structure, and gray fissured bark create distinctive canopy patterns that are clearly visible in the high resolution colour images. Phenologically, teak is semi-deciduous, undergoing a marked seasonal cycle characterized by reddish young leaves during leaf flush, dark green mature leaves, and yellowing and dropping leaves during the dry season \cite{palakit2019influence}. In comparison, many other tropical hardwood trees such as Dipterocarps, Shorea robusta (Sal) and Mahogany (Swietenia) tend to be evergreen with smaller leaves and denser canopies, maintaining stable colors and making them less distinguishable in optical imagery. Therefore, teak’s combination of large leaf morphology, open canopy, visible bark, and seasonal color variation provides distinguishable spectral and textural signals in drone RGB data.

\section{Data Preparation and Methodology}
\subsection{Data Preparation and Annotation}
The methodology begins with the data preprocessing phase, which establishes the foundation for analyzing the 2cm high-resolution images. Teak crown delineation was carried out through manual interpretation (based on in-situ observations) and digitization in QGIS software, with the objective of producing less ambiguous and representative labels. Independent experts manually delineate the tree crowns using QGIS annotation tool. For regions which frequently occur where teak branches heavily interlock with other trees or shadows obscure the edges, a consensus review is introduced to resolve the discrepancies, ensuring accurate and unified ground truth mask before the data is used to train the models. All annotations were created as a binary scheme (teak vs. background) by visually interpreting tree crowns in the imagery and outlining only canopy regions that could be attributed with high confidence. To capture realistic variability in crown appearance, teak was annotated across multiple landscape contexts, including plantation, teak occurring within natural forest matrices, and in deciduous forest patches. This sampling strategy was intentionally designed to incorporate heterogeneity in crown texture, background composition, thereby reducing overfitting to any single forest type and improving model generalization. The teak crowns were labeled as teak when the foliage condition supported unambiguous identification, including trees with young leaves and fully mature leaves. Moreover, crowns exhibiting moderate blurring primarily attributable to wind-induced motion blur were retained within the teak class when the crown boundary and overall teak signature remained interpretable. 

Teak canopies in senescence and leaf-off (canopy absent) conditions were excluded from the teak class due to annotation ambiguity during model training. These conditions may reduce separability from non-teak crowns or background, and therefore were left unlabeled. The following rules were applied to maintain label consistency and avoid artifacts in the training data. Large inter-tree gaps and internal canopy voids were not delineated as teak, and only contiguous canopy surfaces with clear crown area were included in the teak polygons. Shadowed regions were also discarded, as shadow may spectral and textural responses and can lead to mis-classification when trained as part of the teak-crown. To handle class imbalance, a common issue in crown mapping where non-teak or teak dominated training data were organized into image patches with an equal number of teak and background samples. This balance was implemented at the patch level to stabilize contributions from both classes.

\subsection{Models}
For teak and non‑teak segmentation, this study evaluates four widely used semantic segmentation architectures, U‑Net \cite{ronneberger2015u}, PSPNet \cite{zhao2017pyramid}, DeepLabV3 \cite{chen2017rethinking} and one transformer based SegFormer \cite{xie2021segformer} model, each using a ResNet‑34 \cite{he2016deep} encoder initialized with ImageNet \cite{deng2009imagenet} weights to balance representation quality and computational cost. In addition to this, we have trained each model from scratch. The U‑Net employs a symmetric upsampling path with skip connections to recover fine scale crown boundaries, PSPNet adds pyramid pooling to better capture multi scale context around tree clusters, and DeepLabV3 relies on atrous convolutions with spatial pyramid pooling to improve sensitivity to object scale. A final $1 \times 1$ convolution followed by a sigmoid activation in each model produces per pixel probabilities for teak and non‑teak.

\subsection{Training Details}
For model training and validation, the processed UAV RGB images are first divided into smaller image patches of size $224 \times 224$ pixels, using an overlap of 64 pixels to ensure that tree crowns near tile boundaries are well represented. In teak canopy detection from drone imagery, where trees often have a diameter of 5-15 meters (for age above 15 years) the patch size influences how well the model understands what a teak tree-canopy looks like compared to other green vegetation. Since drone imagery has a ground sampling distance (GSD) of 2 cm/pixel, a smaller patch size only covers less than roughly $1.3 \times 1.3$ meters. A mature teak canopy is often much larger than this, which is the reason to create training patches of size $224 \times 224$ pixels. 

The final data set comprises 6391 patches after careful consideration of diverse plots, where samples are randomly taken for training, validation, and testing with an $80:10:10$ ratio, providing a reasonably varying sample set to learn and monitor the performance of the teak and non-teak segmentation model. In addition to this, all image patches are normalized using the min-max normalization in case of training from scratch and when using pretrained weights, standard ImageNet mean and standard deviation values so that the inputs are consistent with the pre-trained ResNet‑34 encoder. To make the model more robust and to expand the effective size of the training set, the images are augmented with horizontal and vertical flips as well as random rotations, capturing different orientations and canopy appearances. The segmentation performance is then evaluated using Intersection over Union (IoU), F1-score and accuracy as the primary metric, providing a clear measure of how well the predicted masks align with the reference annotations.

The model is trained on a system with an NVIDIA Tesla V100 GPU with a RAM of 32 GB. A batch size of 16 is used, and the network is optimized for 100 epochs with an initial learning rate of $1 \times 10^{-3}$, providing enough iterations for the model to converge without overfitting. The training objective combines Dice loss and Binary Cross‑Entropy loss, allowing the network to better cope with class imbalance while still learning well‑calibrated probabilities for the teak and non‑teak classes. All implementations have been carried out using the Segmentation Models Pytorch \cite{Iakubovskii:2019} library.

\begin{table*}[!htb]
\centering
\caption{Quantitative results for native resolution (@2cm) images with patch size of 224 $\times$ 224. Values represents as mean of three different run with $\scriptscriptstyle\pm$ standard deviation.}
\label{tab:model-comparison}
\label{tab:table1}
\resizebox{0.95\textwidth}{!}{

\begin{tabular}{llcccccccc}
\toprule
\multirow{3}{*}{\textbf{Model}} & 
\multirow{3}{*}{\textbf{Pre-Training}} & 
\multirow{3}{*}{\textbf{Params(M)}} & 
\multirow{3}{*}{\textbf{MACs(G)}} & 
\multicolumn{2}{c}{\textbf{F1}} & 
\multicolumn{2}{c}{\textbf{IoU}} & 
\multicolumn{2}{c}{\textbf{Accuracy}} \\
\cmidrule(lr){5-6} \cmidrule(lr){7-8} \cmidrule(lr){9-10}
 &  &  &  & Teak-Tree & All & Teak-Tree & All & Teak-Tree & All \\
\midrule
\midrule
U-Net & \multirow{4}{*}{\textbf{None}} & 24.44 & 6.01 & 0.923$\scriptscriptstyle\pm$0.010 & 0.950$\scriptscriptstyle\pm$0.028 & 0.960$\scriptscriptstyle\pm$0.006 & 0.974$\scriptscriptstyle\pm$0.015 & 0.961$\scriptscriptstyle\pm$0.015 & 0.982$\scriptscriptstyle\pm$0.002   \\
PSPNet &  & 21.44 & 1.18 & 0.879$\scriptscriptstyle\pm$0.031 & 0.921$\scriptscriptstyle\pm$0.048 & 0.935$\scriptscriptstyle\pm$0.018 & 0.958$\scriptscriptstyle\pm$0.026 & 0.953$\scriptscriptstyle\pm$0.004 & 0.970$\scriptscriptstyle\pm$0.009  \\
DeepLabV3&  & 26.01 & 20.94  & 0.914$\scriptscriptstyle\pm$0.013 & 0.944$\scriptscriptstyle\pm$0.032 & 0.955$\scriptscriptstyle\pm$0.007 & 0.971$\scriptscriptstyle\pm$0.017 & 0.948$\scriptscriptstyle\pm$0.007 & 0.980$\scriptscriptstyle\pm$0.003  \\
SegFormer &  & 21.88 & 5.02 & \textbf{0.933$\scriptscriptstyle\pm$0.006} & \textbf{0.956$\scriptscriptstyle\pm$0.004} & \textbf{0.965$\scriptscriptstyle\pm$0.003} & \textbf{0.978$\scriptscriptstyle\pm$0.002} & \textbf{0.967$\scriptscriptstyle\pm$0.004} & \textbf{0.984$\scriptscriptstyle\pm$0.002}  \\

\midrule

U-Net & \multirow{4}{*}{\textbf{ImageNet}} & 24.44 & 6.01 & 0.936$\scriptscriptstyle\pm$0.006 & 0.959$\scriptscriptstyle\pm$0.023 & 0.967$\scriptscriptstyle\pm$0.003  & 0.979$\scriptscriptstyle\pm$0.012 & 0.969$\scriptscriptstyle\pm$0.002 & 0.985$\scriptscriptstyle\pm$0.001  \\
PSPNet &  & 21.44 & 1.18 & 0.929$\scriptscriptstyle\pm$0.002 & 0.954$\scriptscriptstyle\pm$0.025 & 0.963$\scriptscriptstyle\pm$0.001 & 0.976$\scriptscriptstyle\pm$0.013 & 0.966$\scriptscriptstyle\pm$0.001 & 0.984$\scriptscriptstyle\pm$0.000 \\
DeepLabV3&  & 26.01 & 20.94 & 0.931$\scriptscriptstyle\pm$0.007 & 0.955$\scriptscriptstyle\pm$0.025 & 0.964$\scriptscriptstyle\pm$0.004 & 0.977$\scriptscriptstyle\pm$0.013 & 0.957$\scriptscriptstyle\pm$0.014 & 0.984$\scriptscriptstyle\pm$0.002  \\
SegFormer &  & 21.88 & 5.02 & \textbf{0.938$\scriptscriptstyle\pm$0.002} & \textbf{0.960$\scriptscriptstyle\pm$0.022} & \textbf{0.968$\scriptscriptstyle\pm$0.001} & \textbf{0.979$\scriptscriptstyle\pm$0.011} & \textbf{0.968$\scriptscriptstyle\pm$0.002} & \textbf{0.986$\scriptscriptstyle\pm$0.001}   \\
\bottomrule

\end{tabular}
}
\end{table*}

\begin{figure*}[!ht]
\centering
\includegraphics[width=0.99\textwidth]{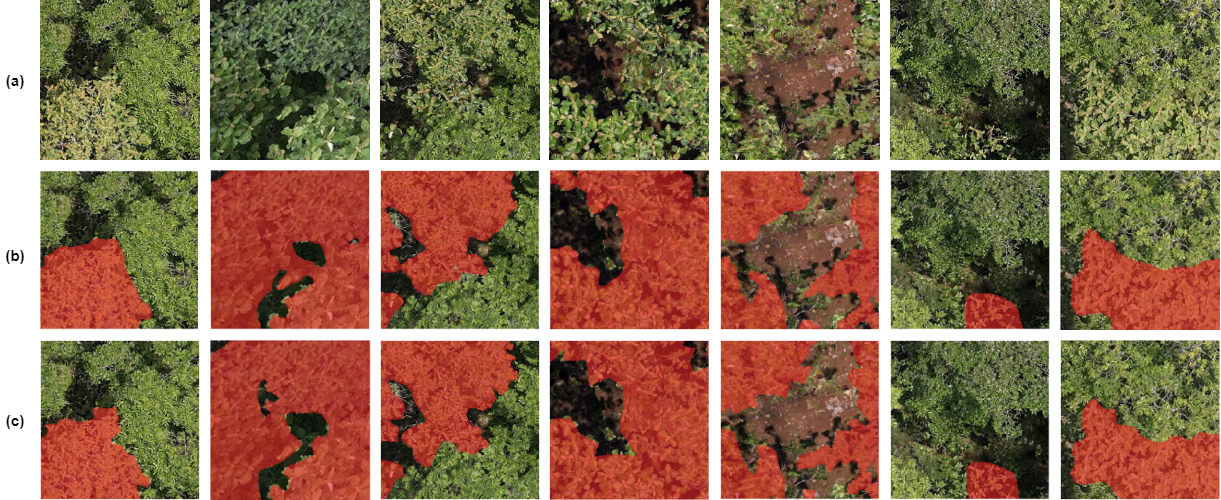}
\caption{Visual comparison of teak/non-teak class segmentation results obtained from different teak canopies. (a) represents original image, (b) shows the generated mask using Segformer model and (c) represents the ground truth}
\label{fig3}
\end{figure*} 
\section{Experiments and Results}
\begin{figure}[!htb]
\centering
\includegraphics[width=0.48\textwidth]{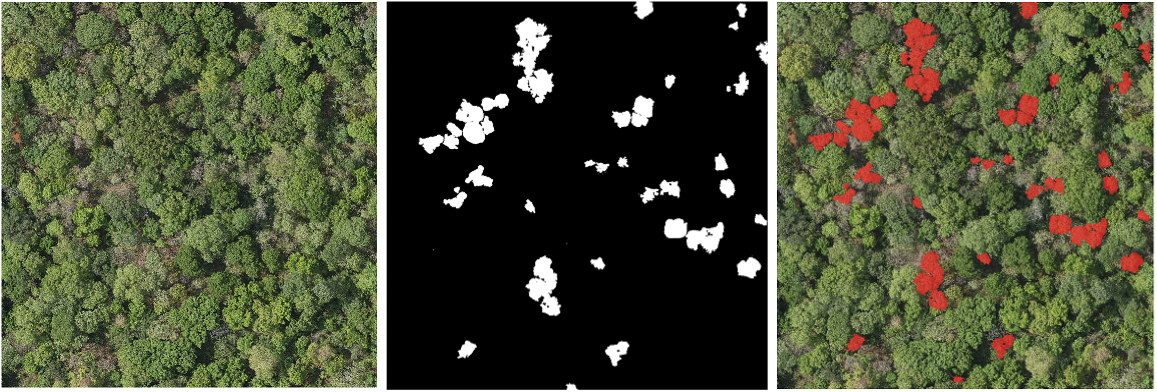}
\caption{Generated full orthomosaic scene. It represents the input full scene, the predicted binary mask and it's overlay over input scene.}
\label{fig5}
\end{figure}

Table~\ref{tab:model-comparison} presents the quantitative evaluation of TeakTreeNet across four state-of-the-art segmentation architectures: U-Net, PSPNet, DeepLabV3, and SegFormer. All models were trained and validated across three independent runs with mean $\pm$standard deviation reported, ensuring statistical robustness. SegFormer consistently outperforms all baselines, achieving the highest F1 (Teak: 0.938$\pm$0.002, All: 0.960$\pm$0.022), IoU (Teak: 0.968$\pm$0.001, All: 0.979$\pm$0.011), and Accuracy (0.986$\pm$0.001) in both scratch and ImageNet-pretrained settings. Notably, SegFormer's computational efficiency (21.88M parameters, 5.02G FLOPs) combined with superior accuracy establishes it as the optimal choice for operational teak plantation monitoring. The visual results for various scenarios have been presented in Figure~\ref{fig3} for Segformer. The baseline approach was qualitatively evaluated across various challenging scenarios, including teak canopies mixed with other tree species, teak canopies against soil backgrounds, teak canopies at different heights resulting in varying leaf sizes, teak canopies with soil and shadow backgrounds, and patches with low teak canopy cover. Visual comparisons presented in Figure~\ref{fig3} demonstrate that the proposed baseline method achieves robust segmentation performance across all these conditions. To obtain a seamless probability map and the final masks for the teak and non-‐teak classes throughout the entire UAV scene, the predictions of the smaller patches are stitched back together taking a average value scheme in the overlapping regions. This mosaicking step helps to smoothly blend adjacent patches, minimizing visible tile boundaries, and producing a spatially coherent segmentation that is more suitable for visualization and downstream forest analysis. Figure~\ref{fig5} represents the full scene mosaic after inference.

\subsection{Impact of Pretrained Weights}
Pretraining with ImageNet weights yields consistent but modest improvements across architectures as shown in Table~\ref{tab:model-comparison}. U-Net benefits most (+1.4\% F1-Teak), while SegFormer shows smallest gains (+0.5\%) despite equivalent parameter counts. ImageNet pretraining reduces variance (average standard deviation drops 45\%), particularly beneficial for PSPNet (F1-All std: from 0.048 to 0.025). However, SegFormer scratch case already outperforms all pretrained CNNs, achieving higher F1-Teak (0.933$\pm$0.006) than PSPNet-ImageNet (0.929$\pm$0.002) with 4.2× fewer FLOPs (5.02G vs 21.18G). This suggests domain-adaptive training suffices for drone-based forestry tasks, avoiding large-scale pretraining overhead. 

\subsection{Cross-region Generalizability}
\begin{figure}[!htb]
\centering
\includegraphics[width=0.48\textwidth]{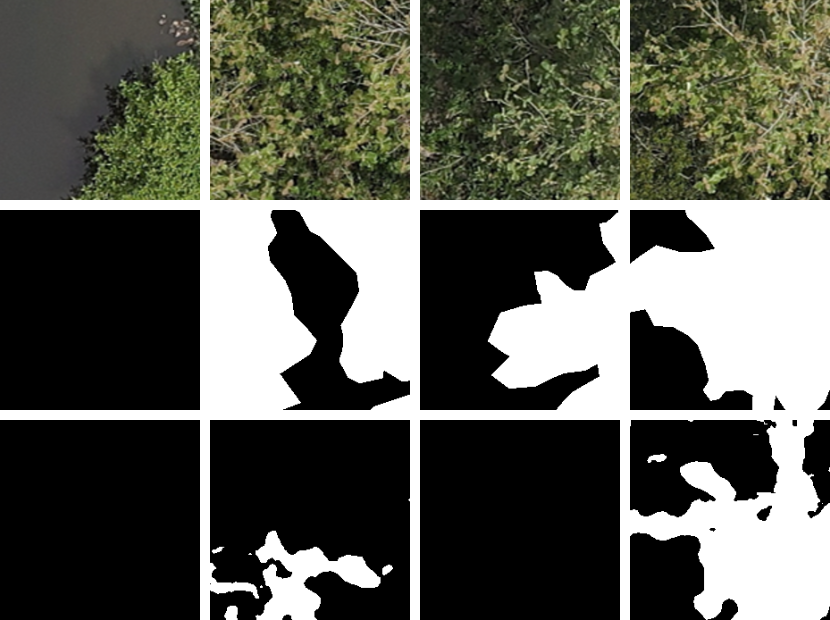}
\caption{Visual results of cross-region validation. First row represents original image, second row represents the ground truth and third row shows the generated mask using pretrained Segformer model}
\label{fig4}
\end{figure}

To evaluate the domain robustness of our teak segmentation model, we trained exclusively on scenes from Shimoga, Karnataka, India and tested on an independent location Nilambur, Kerala, India without considering samples from Nilambur while training (patches with low teak canopy cover during the leaf-fall phenological phase, 2023). Under this cross‑region setting, the top performing model i.e SegFormer attains a mean IoU of 0.574 (IoU = 0.962 for non-teak canopy and IoU = 0.186 for teak canopy). This large inter‑class gap indicates that the model retains its ability to recognize background land cover but fails to generalize teak detection to the new domain, likely due to domain shift in spectral signatures, canopy texture, plantation structure, contextual cues between locations and difference in teak phenology in both the locations. These results highlight the limitations of single‑region training for operational forest monitoring and underscore the need for multi‑region training, domain adaptation, or targeted fine‑tuning on a small amount of labeled data from new regions to achieve robust, transferable teak mapping at scale. For instance, The cases shown in Figure \ref{fig4} primarily correspond to patches with low teak canopy cover during the early leaf-emergence phenological phase, where sparse teak foliage exposes the background non-teak vegetation canopy. Such conditions increase the complexity of canopy discrimination and may lead to segmentation errors in detecting teak canopies. The results suggest that dense, fully developed teak canopies yield more accurate and robust detection performance than sparsely foliated canopies.

\section{Conclusion}
This work establishes TeakTreeNet as the benchmark for teak and non-teak tree binary segmentation, addressing a critical gap in drone-based forestry remote sensing. Our comprehensive evaluation of four state-of-the-art architectures reveals SegFormer's better performance (F1 0.938±0.002, IoU 0.968±0.001, Accuracy 0.986±0.001) with minimal computational overhead (21.88M params, 5.02G FLOPs), outperforming ImageNet-pretrained CNNs across all metrics. The observations of this study include the domain-adaptive training from scratch in which rivals transfer learning for specialized remote sensing tasks, enabling efficient deployment without pretraining infrastructure. Moreover, The primary limitation of the proposed approach is its reduced performance in patches with sparse teak canopy cover during early phenological stages, where substantial exposure of background vegetation increases spectral and structural ambiguity. Therefore, image acquisition should preferably be conducted during periods when teak canopies are fully developed, as dense foliage provides more distinctive canopy characteristics and enables more accurate and reliable detection than sparsely foliated canopies. The incorporation of multi-spectral and multi-temporal datasets in future studies may improve the model's ability to account for phenological variations and ensure more consistent segmentation throughout the year.




%



\section*{Acknowledgment}
The authors gratefully acknowledge the financial support provided by the NASA-ISRO Synthetic Aperture Radar (NISAR) Science Programme of the Indian Space Research Organisation (ISRO) for carrying out this research.

\ifCLASSOPTIONcaptionsoff
  \newpage
\fi



\bibliographystyle{IEEEtran}
\bibliography{IEEEabrv,main}
%

\end{document}